\documentclass[10pt,twocolumn,letterpaper]{article}

\usepackage{cvm}
\usepackage{times}
\usepackage{epsfig}
\usepackage{graphicx}
\usepackage{amsmath}
\usepackage{amssymb}
\usepackage{color}

\usepackage[pagebackref=true,breaklinks=true,letterpaper=true,colorlinks,bookmarks=false]{hyperref}
\usepackage[capitalise, nameinlink]{cleveref}

\crefname{section}{Sec.}{Sec.}

\newcommand{\keywords}[1]{{\bf \emph{Keywords: #1}}}
\newcommand{\warning}[1]{{\color{black}#1}}

\cvmfinalcopy
\def\cvmPaperID{****} % *** Enter the cvm Paper ID here

\ifcvmfinal\fi
\begin{document}

%%%%%%%%% TITLE
\title{A Real-time Method for Inserting Virtual Objects into Neural Radiance Fields}
%\title{Inserting Virtual Objects into Neural Radiance Fields}

\author{Keyang Ye\textsuperscript{1}, Hongzhi Wu\textsuperscript{1}, Xin Tong\textsuperscript{2}, Kun Zhou\textsuperscript{1}\\
\textsuperscript{1} State Key Lab of CAD\&CG, Zhejiang University, Hangzhou, China\\
\textsuperscript{2} Microsoft Research Asia, Beijing, China \\
{\tt\small kunzhou@acm.org}
% \and
% {\small\url{http://www.author.org/~second}}
}

\maketitle
% \thispagestyle{empty}

%%%%%%%%% ABSTRACT
\begin{abstract}
    We present the first real-time method for inserting a rigid virtual object into a neural radiance field, which produces realistic lighting and shadowing effects, as well as allows interactive manipulation of the object. By exploiting the rich information about lighting and geometry in a NeRF, our method overcomes several challenges of object insertion in augmented reality. For lighting estimation, we produce accurate, robust and 3D spatially-varying incident lighting that combines the near-field lighting from NeRF and an environment lighting to account for sources not covered by the NeRF. For occlusion, we blend the rendered virtual object with the background scene using an opacity map integrated from the NeRF. For shadows, with a precomputed field of spherical signed distance field, we query the visibility term for any point around the virtual object, and cast soft, detailed shadows onto 3D surfaces. Compared with state-of-the-art techniques, our approach can insert virtual object into scenes with superior fidelity, and has a great potential to be further applied to augmented reality systems.
\end{abstract}


%%%%%%%%% BODY TEXT
\section{Introduction}
Neural Radiance Field (NeRF) is a popular technique for novel view synthesis, by representing a scene with implicit fields of density and view-dependent color, trained end-to-end with respect to input images~\cite{original_nerf}. Substantial research efforts have been made to extend the original work to different scenarios (e.g., dynamic scenes~\cite{dynamic_nerf}, human bodies~\cite{nerf_human_act} or illumination variations~\cite{nerf_w}).

Despite that NeRF is a promising, novel representation with rich information about a scene, its application to augmented reality (AR) is still limited. Related work focuses on inserting a NeRF into another one~\cite{editable_nerf}, or a NeRF onto a background photograph~\cite{cao2023real}. To our knowledge, very few existing techniques exploit the information in a NeRF to perform virtual object insertion, a classic AR task.

\warning{While our task looks straightforward at first glance, a number of challenges arise in inserting a virtual object into a NeRF. First, the input NeRF may not completely cover the complete lighting information of a scene. How to compensate for the potentially missing light sources? Second, it is difficult to represent and/or precompute near-field lighting in an efficient manner. Moreover, complex occlusion/shadowing effects between the virtual object and the NeRF must be modeled with high fidelity and performance.

%The key to this common task is the proper estimation of lighting and geometry from the NeRF. Without accurately estimated lighting, the virtual objects would be too bright/dark, or exhibit unconvincing reflections. On the other hand, without accurate scene geometry, the scene could not correctly occlude the virtual object, and the shadow cast by the virtual object would be problematic.
To tackle the above challenges, we present in this paper the first real-time method for inserting a virtual object into a neural radiance field, which produces realistic rendering with shadowing and occlusion effects, as well as allows interactive manipulation of the virtual object. Our method takes as input an HDR NeRF and one or more virtual objects only, and requires a modest overhead for precomputation and storage. To account for light sources not covered by NeRF, we estimate a distant environment lighting with inverse rendering. To efficiently handle near-field lighting, we sample the NeRF and combine with the environment lighting as the incident lighting, expressed in spherical Gaussians (SG) for high-performance rendering. To rapidly model shadowing effects, we precompute the self-visibility of the virtual object, and store the result as a field of spherical signed distance fields (SSDF)~\cite{all_freq}, which can be queried efficiently in real time.} Our method compares favorably with state-of-the-art techniques in generating high-fidelity insertion results.

Our main contribution is the first complete framework to insert a virtual object into a neural radiance field in real time, an increasingly popular representation. The rest of this paper is structured as follows. \cref{related} discusses the related work, and~\cref{methods steps} describes our method. In~\cref{experiment sec}, we present experiment results for both synthetic and real scenes. Finally, \cref{conclusion sec} concludes the paper.

\section{Related Work}
\label{related}
%Our approach lies at the intersection of multiple fields. We briefly review the related prior works below.
Below we review three categories of work most related to this paper.

%Lighting estimation is a classic challenge in computer graphics. It is critical for realistic relighting in objects insertion. We categorize the previous methods into scene-based and object-based methods. 
%and the input image is considered as a partial content of the environment map, so it is similar to the task of image completion. 

\textbf{Lighting estimation.} Here previous work can be divided into scene-based and object-based methods. Scene-based methods estimate the lighting from partial view(s) of the scene, which is similar to the task of image completion. The missing part is approximated by directly copying from input images~\cite{khan2006image}, or by searching a panorama database for an environment map similar to the input~\cite{karsch2014automatic}. Recently, deep learning is employed to fill in the missing information. Convolutional Neural Network (CNN) or Generative Adversarial Network (GAN) are used to predict an environment map from input images with limited fields of view~\cite{gardner2017learning, deeplight, emlight, ma2021neural}. Garon et al.~\cite{garon2019fast} take images with coordinate masks and corresponding local patches as input to a neural network to predict 2D spatially-varying lighting expressed in spherical harmonics (SH). However, learning-based methods suffer from weak generalization ability, when the input images differ considerably from the training set. For example, the models trained on indoor scene datasets are difficult to generalize to outdoor~\cite{emlight, deep_parametric}.

On the other hand, object-based methods estimate lighting from the appearance of object(s) presented in the scene. A common technique is to jointly optimize the object material and the environment lighting by inverse rendering~\cite{nerfactor, nerv, nerd, physg, zhang2022modeling}. This is a highly ill-posed problem, and its solution often requires strong priors, such as smoothness~\cite{nerfactor} or Lambertian only~\cite{azinovic2019inverse}. Recently, deep learning has also been applied. Much of related work is targeted for human faces, due to the existing prior knowledge about their shapes and appearance~\cite{tewari2017mofa, calian2018faces}.

Our method applies inverse rendering to estimate a distant environment lighting from a NeRF to compensate for potential missing lights not recorded in the NeRF. We take both the near-field lighting from NeRF and the distant environment lighting into consideration and build an inverse rendering pipeline, similar to~\cite{zhang2022modeling}.

%Therefore, different from works that focus on object-centric scenes and consider objects are only illuminated by environment lighting~\cite{physg, nerfactor}. Our method takes both the local lighting in NeRF and environment lighting from infinity into consideration and build a inverse rendering pipeline based on the method of Zhang et al~\cite{zhang2022modeling}.

\textbf{Visibility estimation.} Most contemporary AR systems estimate depths for visibility computation. Depths can be directly acquired with specialized devices~\cite{kendall2017end, bhat2021adabins, chang2018pyramid}. To reduce the hardware requirement, single-view approaches apply deep learning~\cite{alhashim2018high, godard2017unsupervised} or employ inverse rendering that adds depths as additional unknowns to a joint optimization~\cite{IRAdobe}. When a sequence of images is available, Multi-View Stereo (MVS) estimates depths by matching pixels~\cite{wang2018mvdepthnet} or features~\cite{mvs_net}. Waston et al.~\cite{watson2023virtual} sidestep the depth estimation and directly predict the occlusion mask from a feature map. However, these methods still fall short in some cases, especially on translucent objects or objects with tiny structures, such as leaves on plants. In comparison, our method exploits the implicit geometry in a NeRF, which can handle cases that are challenging for traditional explicit representations.

\textbf{Real-time shadows.} Shadow field~\cite{shadow_field} precomputes visibility maps of samples from concentric shells surrounding the virtual object and use SH as basis to compress them. However, the approach only supports soft shadows when using low order SH basis. Wang et al.~\cite{all_freq} introduce spherical signed distance fields (SSDF) for high-frequency shadows. Their method precomputes the SSDF at every vertex that receives shadows, and compresses them with PCA method, which requires both the scene and the object being static. Kei et al.~\cite{all_freq_sphere} approximate objects with a set of spheres and project them to the hemisphere integral region of the shading point as occluded patches. Dynamic objects are supported at the cost of intensive computation. We propose an SSDF field, which combines the idea of shadow field and SSDF, to efficiently render shadows cast by the virtual object.

\section{Preliminaries}
\label{preliminary}
A NeRF represents a scene with a volumetric field $\mathcal{F}$ of density $\sigma$, and color $\mathbf{c}$ which varies with a view direction $\mathbf{d}$. For a ray $\mathbf{r}(t) = \mathbf{o}+\mathbf{d}t$ emitted from camera center $\mathbf{o}$ along $\mathbf{d}$, the rendered color $\mathbf{\hat{C}}(\mathbf{d}; \mathbf{o})$ of a NeRF is computed with simple volumetric rendering~\cite{original_nerf}.
The opacity $\hat{O}(\mathbf{d}; \mathbf{o})$ and depth $\hat{D}(\mathbf{d}; \mathbf{o})$ are computed in a similar fashion: $\hat{O}(\mathbf{d}; \mathbf{o}) = \sum_{k=1}^K T\left(t_k\right) \alpha\left(\sigma\left(t_k\right) \delta_k\right)$ and $\hat{D}(\mathbf{d}; \mathbf{o}) = \sum_{k=1}^K T\left(t_k\right) \alpha\left(\sigma\left(t_k\right) \delta_k\right) t_k$. Here $t_k$ is a distance of a point sample traveled along the ray, $t_\mathrm{k}\in[t_\mathrm{n}, t_\mathrm{f}]$, where $t_\mathrm{n}$ and $t_\mathrm{f}$ are the near and far distances of the intersection of the NeRF boundary and the ray. Please refer to~\cref{sample} for an illustration of $t_\mathrm{k}$. $T\left(t_k\right)$ is the cumulative transmittance, $\sigma(t_k)$ are the density at the sample point on the ray $\mathbf{r}(t)$. In addition, $\alpha(x) = 1-e^{-x}$ and $\delta_k = t_{k+1} - t_k$ is the distance between adjacent two samples along the ray. 

NeRF can be trained from a set of images with known intrinsic and extrinsic camera parameters. While low-dynamic-range (LDR) images are usually used, they are not suitable for physically based rendering. Therefore, we use high-dynamic-range (HDR) NeRF as input to our pipeline throughout the paper. Moreover, we employ instant-ngp~\cite{instant_ngp} as the implementation, due to its excellent performance.

%\subsection{Assumptions}
%\label{assumption}
%Our target is to insert a virtual object into a NeRF in real time. To this end, we make some assumptions for the virtual object and the scene. 

\textbf{Assumptions.} We assume that the virtual object is rigid and opaque. Also its size is relatively small with respect to the input NeRF, so that the incident lighting (ignoring occlusion) does not change over its surface. For the input NeRF, we assume that the geometry incorporated in it can be sampled as surface points, and its materials are Lambertian. In the whole pipeline, we only consider the direct illumination and ignore interreflection effects.

\warning{Note that the above assumptions may be loosened, at the expense of more runtime computations. For example, one may build the incident lighting at sampled locations across the object surface, and interpolate them to obtain the result for any point. For the surface assumption of the NeRF, one can remove it by switching to a more involved inverse rendering computation, which takes scattering into account. On the high level, our framework is not limited to the above assumptions and can be extended to handle more diverse cases.}

\section{Our method}
\label{methods steps}

\subsection{Overview}
\label{overview}

\begin{figure*}[h]
\centering
\includegraphics[width=1.0\textwidth]{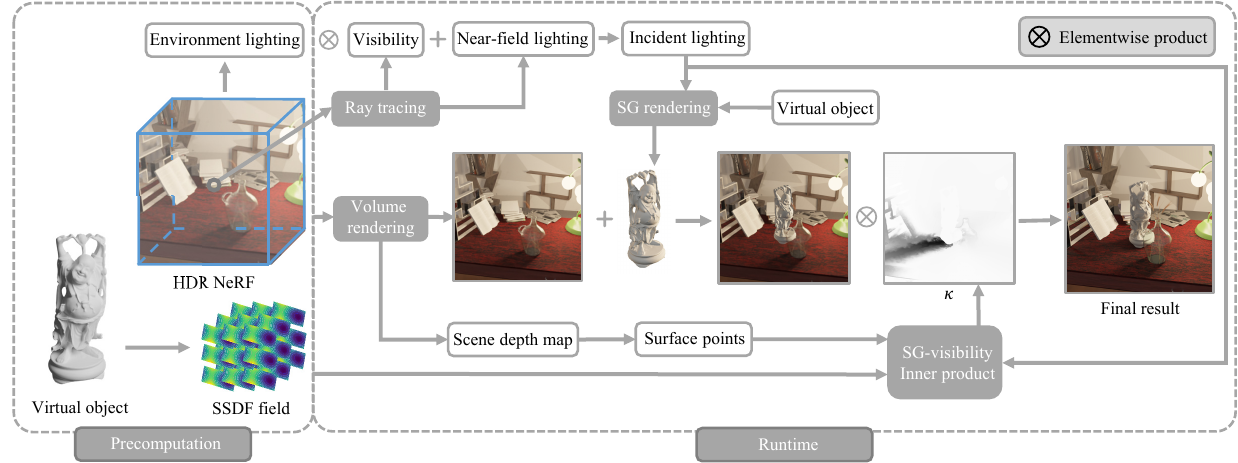}
\caption{The full pipeline of our method. Our method takes an HDR NeRF and a rigid virtual object as input. We estimate the environment lighting to compensate for potential lighting not covered by the NeRF. Next, we precompute the visibility field of the virtual object as a SSDF field. At runtime, we perform ray tracing from the virtual object center to obtain the near-field lighting and the visibility of NeRF, and compute the incident lighting fitted with spherical Gaussians (SG) to fast render the virtual object. The rendered virtual object will be composited into the NeRF by blending. We also use the incident light and the SSDF field to compute $\kappa$ for shadows at each pixel (\cref{shadow ratio}), and multiply the pixel color by the corresponding $\kappa$ to produce the final result.} 
\label{fig_pipeline}
\end{figure*}

%We summarize our method into five steps, including high dynamic range (HDR) NeRF training, lighting estimation, virtual object rendering, virtual object compositing and shadow casting. Our method takes multi-view HDR images, an opaque virtual object (meshes) and its materials (albedo, roughness and metallic) and transformation (rotation, translation and scaling), camera poses and some precomputation data as inputs, and outputs object insertion results as HDR images. The full pipeline of our method is shown in~\cref{fig_pipeline}. We use different colors to denote the corresponding sections of the pipeline.
We take as input an HDR NeRF (denoted as $\mathcal{F}$) and a rigid virtual object with geometry (represented as a mesh) and appearance (parameters stored in texture maps). First, to compensate for potential lighting not covered by the NeRF, we extrapolate it to an additional environment lighting, and then combine both into an incident lighting expressed in SGs for fast evaluations (\cref{env light est sec}). Next, to facilitate rapid shadow computation, we precompute the visibility field around the virtual object as a field of SSDF (\cref{SSDF fields sec}). At runtime, we render the virtual object with estimated incident lighting and the visibility field, and composite with the original NeRF as the output in real time (\cref{object insertion sec}). A graphical illustration of the pipeline is shown in~\cref{fig_pipeline}.

\warning{Note that an alternative option is to precompute an field of incident lighting, expressed in SGs. However, the non-linear nature of SGs makes the interpolation difficult, which motivates our current approach for lighting estimation.}

\subsection{Environment lighting estimation}
\label{env light est sec}
Ideally, all incident lighting of the virtual object comes from $\mathcal{F}$. In practice, however, 
$\mathcal{F}$ may not cover the complete surroundings: there might be light sources in the scene, which are not recorded by $\mathcal{F}$. To alleviate this issue, we extrapolate $\mathcal{F}$ to an additional distant environment lighting to fully cover the virtual object. Specifically, we formulate the incident lighting at a surface point $\mathbf{x}$ of the virtual object along a direction $\mathbf{d}$ as follows:

\begin{equation}
L_\mathrm{i}(\mathbf{d}; \mathbf{x}) = (1-\hat{O}(\mathbf{d}; \mathbf{x})) L_\mathrm{env}(\mathbf{d}) + \hat{O}(\mathbf{d}; \mathbf{x}) L_\mathrm{nerf}(\mathbf{d}; \mathbf{x}),
\label{incident_light}
\end{equation}
%$\hat{O}(\mathbf{d}; \mathbf{x})$
where $\hat{O}(\mathbf{d}; \mathbf{x})$ is the NeRF opacity according to~\cref{preliminary}, and $L_\mathrm{i}$, $L_\mathrm{env}$, $L_\mathrm{nerf}$ are the incident lighting, the distant environment lighting and the near-field lighting from $\mathcal{F}$, respectively.

%The NeRF visibility is 0 when rays intersect with NeRF, otherwise it is 1. When there are translucent objects in the scene, the visibility value may be between 0 and 1. We use the opacity in~\cref{opac_eq} to compute the NeRF visibility as $V(\mathbf{d}; \mathbf{x}) = 1-\hat{O}(\mathbf{d}; \mathbf{x})$. 
%The computation model of the ambient light is illustrated in light red area of~\cref{fig_pipeline}. The local lighting, including indirect (light yellow arrow) and direct (yellow arrows) light, can be computed by emitting sample rays from the object center to all directions and integrating color results using~\cref{nerf_integral}. With the use of the local lighting, indirect lighting and local occlusion of lights can be modeled, which are often ignored by other methods~\cite{physg}.

%\begin{figure}[h]
%\centering
%\includegraphics{imgs/第6页.pdf}
%\caption{Computation of ambient light. The ambient light is computed at the center of the virtual object, and it consists of local light and global light. The local light, including local indirect and direct light, can be obtained from NeRF. Light sources outside the scene bounding box are considered as global light from infinity.}
%\label{global_local_light}
%\end{figure}

For the environment lighting, we represent it as the sum of $M=32$ spherical Gaussian (SG) lobes, which strike a good balance between the fidelity and rendering efficiency~\cite{all_freq}:
\begin{equation}
L_\mathrm{env}(\mathbf{d}) = \sum_{k=1}^M G(\mathbf{d}; \mathbf{p}_k, \lambda_k, \boldsymbol{\mu}_k),
\label{SG_light}
\end{equation}
where $\mathbf{p}_k\in\mathbb{S}^2$ is the center, $\lambda_k\in\mathbb{R}^+$ is the sharpness, $\boldsymbol{\mu}_k\in\mathbb{R}_+^n$ is the amplitude, for a particular SG. 

%\warning{view sampling of NeRF. Specifically, we randomly sample camera positions on the sphere with the NeRF center as its center and half of the NeRF grid size as its radius. We let all cameras look at the NeRF center and render views. To remove views of poor quality (usually the views outside the range of NeRF training), we score the views by the no-reference image quality assessment method: BRISQUE~\cite{brisque}, and save the views with the score lower than 50 (the higher the quality, the lower the score). We repeatedly perform the sampling until there are 100 views available.} 
To estimate the SG parameters of an environment lighting, we first randomly sample views that pointing towards the center of the NeRF volume, with a distance of half of the NeRF grid size. The view will be rejected if the image quality of the corresponding NeRF rendering result is not sufficient (BRISQUE~\cite{brisque} score $>$ 50). We repeat the process until we have 100 sampled views. For the rendered image at each sampled view, we randomly sample 64 pixels that pass through $\mathcal{F}$. For each such pixel, we compute a surface point sample as: $\mathbf{x}_\mathrm{surf} = \mathbf{o} + \hat{D}(\mathbf{d}; \mathbf{o})\mathbf{d}$, similar to~\cite{nerfactor}, where $\mathbf{o}$ is the camera center. The normal of this sample is estimated by the gradient of density as $\mathbf{n}(\mathbf{x}_\mathrm{surf}) = -\nabla\sigma(\mathbf{x}_\mathrm{surf})$. Finally, the environment lighting is linked with the rendered pixels ($L_\mathrm{o}(\mathbf{x})$) according to the rendering equation as follows, which allows us to perform inverse rendering to optimize the SG parameters:
\begin{equation}
    L_\mathrm{o}(\mathbf{x}) = \frac{\mathbf{a}(\mathbf{x}; \mathbf{\Theta_\mathrm{a}})}{\pi}(E_\mathrm{nerf}(\mathbf{x})+E_\mathrm{env}(\mathbf{x}; \mathbf{\Theta_\mathrm{env}})).
\end{equation}  
Here $\mathbf{a}$ is the albedo, represented as a 6-layer MLP~\cite{zhang2022modeling} (whose parameters are $\mathbf{\Theta_\mathrm{a}}$) and jointly optimized along with the environment lighting. $\mathbf{\Theta_\mathrm{env}}$ are the SG parameters. $E_\mathrm{nerf}$ is the irradiance integrated from $\mathcal{F}$ via Monte-Carlo sampling, which is only computed once throughout the optimization. For $E_\mathrm{env}$, we first cache the NeRF opacity at each sample point into texture maps, and then perform importance sample according to the mixture of SGs to compute the integral during the optimization, similar to the method of Zhang et al.~\cite{zhang2022modeling}.

\subsection{Visibility field precomputation}
\label{SSDF fields sec}
Once the environment lighting is estimated, the next step is to precompute the visibility for any point around the virtual object, so that such information can be rapidly retrieved during runtime for shadow computation. Inspired by shadow field~\cite{shadow_field}, we precompute a field of SSDF around the virtual object, which allows fast rendering with incident lighting expressed as SGs~\cite{all_freq}.

%. Note that while a more natural and consistent choice to express precomputed visibility is via a field of SGs, its non-linear nature makes interpolation difficult. Therefore, we adopt SSDF~\cite{all_freq}, which also allows fast rendering with incident lighting expressed as SGs~\cite{all_freq}.}

First, a spherical signed distance field at a point $\mathbf{x}$ around the virtual object is defined as follows. Given a ray from $\mathbf{x}$ along the direction $\mathbf{d}$, the SSDF $S(\mathbf{d}; \mathbf{x})$ is the minimal angle between $\mathbf{d}$ and the direction from $\mathbf{x}$ to the the silhouette of the virtual object. The angle is positive when $\mathbf{d}$ does not intersect with the object, and negative otherwise. Interested readers are directed to~\cite{all_freq} for a detailed derivation.

Next, we precompute the SSDFs at sampled points around the virtual object. Similar to~\cite{shadow_field}, we use the points on a uniform 3D grid of $16\times16\times16$, whose center coincides with that of the object $\mathbf{x}_\mathrm{o}$. The length of the grid is 3 times the radius of the bounding sphere of the object. Furthermore, we perform principal component analysis (PCA) to compress all sampled SSDFs to 1.8\% of the original size.

At runtime, the precomputed SSDF samples are processed to a continuous field for pixel-level rendering.
For a point inside the grid, the SSDF can be obtained via a trilinear interpolation of the SSDFs at related sampled points. For a point $\mathbf{x}_\mathrm{f}$ outside the grid, we snap to $\mathbf{x}_\mathrm{b}$, the nearest point in the grid. We adjust the SSDF at $\mathbf{x}_\mathrm{b}$ with simple geometric relations to approximate that of $\mathbf{x}_\mathrm{f}$.

\subsection{Real-time virtual object insertion}
\label{object insertion sec}
With the precomputed SSDF field, we perform virtual object insertion as follows. We compute the incident lighting by ray-tracing at the center of the virtual object according to~\cref{env light est sec} and fit it with SGs \warning{(which we refer to it as SG updating)}, render the object, composite it into the NeRF and add shadows at each frame. Note that for SG fitting, we adopt the estimates from the previous frame as initial values, both for temporal stability and faster convergence. Below are detailed descriptions.

%The latter three steps are detailed below.

%\warning{Note that for runtime performance, the above SG optimization takes the estimates at the previous frame as initial values, which we refer to as SG updating in the remainder of the text.}

%while the amplitude of SGs are multiplied by different self-shadow ratios $\gamma$ at different surface points to yield the self occlusion effects. We will explain $\gamma$ in the shadow effects part.
%We use the incident light at the virtual object center to render the entire object, according to our assumption in~\cref{preliminary}. Specifically, we first perform ray tracing to obtain the incident light by~\cref{incident_light} and use 32 SG lobes to fit it. Note that SG lobes do not form a set of orthonormal bases, which means that we must use numerically optimization method to perform fitting. Therefore, we randomly initialize the parameters of SG lobes at the first frame. For subsequent frames, we use the optimization results of the previous frame as the initialization, which preserves temporal stability and accelerates convergence when the virtual object moves slowly. The incident light at each surface point of the virtual object is approximated by the incident light at the virtual object center,
\textbf{Virtual object rendering}. To model the object appearance, we adopt the simplified Disney BRDF model~\cite{disney_brdf} with roughness $r$, metallic $m$ and diffuse albedo $\mathbf{a}$, and assume a fixed $F_0=0.02$ in the Fresnel term. As in previous work~\cite{physg, all_freq}, the BRDF $f_r$ and the cosine term $\left(\boldsymbol{\omega}_\mathrm{i} \cdot \mathbf{n}\right)$ can be converted to mixtures of SGs as well. Therefore, the rendering computation is approximated as a rapid inner product of SGs using a fragment shader:
\begin{equation}
L_\mathrm{o}\left(\boldsymbol{\omega}_\mathrm{o} ; \mathbf{x}\right) = \sum_{k=1}^M (G(\boldsymbol{\omega}_\mathrm{i}; \mathbf{p}_k, \lambda_k, \gamma_k(\mathbf{x})\boldsymbol{\mu}_k) \otimes f_\mathbf{r}) \cdot \left(\boldsymbol{\omega}_\mathrm{i} \cdot \mathbf{n}\right),
\label{SG_render_eq}
\end{equation}
where $\mathbf{p}_k, \lambda_k$ and $\boldsymbol{\mu}_k$ are the SG parameters of the incident lighting at the virtual object center. $\gamma_k$ attenuates the final result to account for self-shadows, which we will detail in the description of shadow computation. $\otimes$ represents direction-wise multiplication (SG product). As the BRDF is represented with SGs, this allows direct editing of appearance in real time.

%The shape of BRDF SG is directly related to BRDF parameters, so we can edit the material in real time without any extra time consumption.

\begin{figure}
\centering
\includegraphics[width=0.37\textwidth]{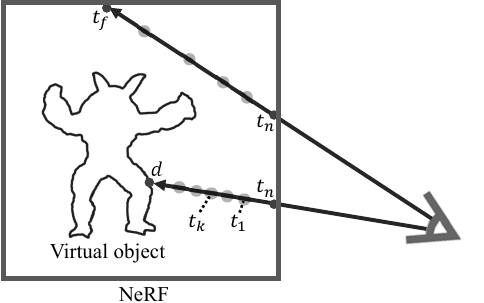}
\caption{The illustration of sampling points for volume rendering. Rays are emitted from the camera center to the NeRF to render the view. For rays not intersecting with the virtual object, we sample points with the distance range $t_\mathrm{k}\in[t_\mathrm{n}, t_\mathrm{f}]$, where $t_\mathrm{n}$ and $t_\mathrm{f}$ are the near and far distances of the intersection of the NeRF boundary and the ray. Otherwise, we sample points with the distance range $t_\mathrm{k}\in[\min (d, t_\mathrm{n}), \min (d, t_\mathrm{f})]$, where $d$ is the depth value of the virtual object, to eliminate the part of the NeRF occluded by the virtual object. }
\label{sample}
\end{figure}

\textbf{Compositing}. Here we mix the rendering results of the input NeRF and the virtual object. First, for a pixel whose camera ray hits the virtual object, we adjust the integration range $t$ to $[\min (d, t_\mathrm{n}), \min (d, t_\mathrm{f})]$, where $d$ is the depth of the object at the current pixel. The idea is to eliminate the part of the NeRF occluded by the virtual object. Please refer to~\cref{sample} for an illustration. We then perform conventional NeRF rendering to obtain a color and an opacity map, $\mathcal{I}_\mathrm{c}$ and $\mathcal{I}_\mathrm{\alpha}$. Next, we blend in the rendering result of the virtual object (without shadows) $\mathcal{I}_{v}$ as:
\begin{equation}
\mathcal{I}_\mathrm{\alpha} \mathcal{I}_\mathrm{c} + (1 - \mathcal{I}_\mathrm{\alpha}) \mathcal{I}_\mathrm{v}.
\label{blend_result}
\end{equation}

%We first perform conventional NeRF rendering to obtain a color and an opacity map, $\mathcal{I}_\mathrm{c}$ and $\mathcal{I}_\mathrm{o}$. Note that the distance range of a point sample in~\cref{preliminary} is now changed to $t_\mathrm{k}\in[\min (d, t_\mathrm{n}), \min (d, t_\mathrm{f})]$, where $d$ is the depth value of the virtual object. We illustrate the sampling method in~\cref{sample}.

% We denote the rendering color and depth map of the virtual object as $\mathcal{I}_\mathrm{vc}$ and $\mathcal{I}_\mathrm{vd}$ respectively. The scene is rendered by emitting rays from the camera center to the image plane and performing volume rendering to obtain color $\mathcal{I}_\mathrm{nc}$ and opacity $\mathcal{I}_\mathrm{no}$ results. Note that the distance range of the sampling ray $\mathbf{r}(t) = \mathbf{o} + t\mathbf{d}$ is now changed to $t\in[\min (d(\mathbf{r}), t_\mathrm{n}), \min (d(\mathbf{r}), t_\mathrm{f})]$, where $d(\mathbf{r})$ is the depth value queried from $\mathcal{I}_\mathrm{vd}$. Finally, we blend the rendering results as follow:
% \begin{equation}
% \mathcal{I}_\mathrm{c} = \mathcal{I}_\mathrm{nc} + \mathcal{I}_\mathrm{vc} \otimes (1 - \mathcal{I}_\mathrm{no}),
% \label{blend_result}
% \end{equation}
% where $\mathcal{I}_{c}$ is the compositing result (without shadows), and $\otimes$ means element-wise multiplication. We achieve detailed opaque and translucent occlusion effects between the scene and the virtual object without using the explicit geometry of the scene, such as the depth map. 

\textbf{Shadow computation}. We handle two types of shadows in our pipeline: the shadow from the virtual object to the NeRF, and the self-shadow of the virtual object. 

For the first type of shadows, similar to previous work~\cite{wang2022neural}, we compute an attenuation factor $\kappa$, as the ratio of the irradiance value before and after the virtual object insertion:
\begin{equation}
\begin{aligned}
\kappa(\mathbf{x})&=\frac{\int_{\Omega^+} L_\mathrm{i}\left(\boldsymbol{\omega}_\mathrm{i}; \mathbf{x}_\mathrm{o}\right) V\left(\boldsymbol{\omega}_\mathrm{i}; \mathbf{x}\right)\left(\boldsymbol{\omega}_\mathrm{i} \cdot \mathbf{n}\right) \mathrm{d} \boldsymbol{\omega}_\mathrm{i}}{\int_{\Omega^+} L_\mathrm{i}\left(\boldsymbol{\omega}_\mathrm{i}; \mathbf{x}_\mathrm{o}\right)\left(\boldsymbol{\omega}_\mathrm{i} \cdot \mathbf{n}\right) \mathrm{d} \boldsymbol{\omega}_\mathrm{i}}.
\end{aligned}
\label{shadow ratio}
\end{equation}
Here $\mathbf{x}_\mathrm{o}$ is the object center. $V$ is the binary visibility term: it is 0 if occluded by the virtual object, and 1 otherwise. For each pixel in $\mathcal{I}_\mathrm{c}$, we compute its corresponding 3D location $\mathbf{x}$ from its depth using the same method in~\cref{env light est sec}, and then attenuate its rendered result by a corresponding $\kappa$.

\cref{shadow ratio} can be rapidly computed thanks to the SG basis. For its numerator, we first multiply $L_\mathrm{i}$ with the cosine term as a product of SGs, and then compute as follows:
\begin{equation}
\begin{aligned}
\int_{\Omega}(\sum_{k=1}^M G(\boldsymbol{\omega}_\mathrm{i}; \mathbf{p}_k, \lambda_k, \boldsymbol{\mu}_k))V(\boldsymbol{\omega}_\mathrm{i}) \mathrm{d}\boldsymbol{\omega}_\mathrm{i} \\
\approx \sum_{k=1}^M\boldsymbol{\mu}_k f_\mathrm{h}(S(\mathbf{p}_k), \lambda_k),
\label{SG_visi_product}
\end{aligned}
\end{equation}
where $f_\mathrm{h}(\theta, \lambda)$ is a precomputed table~\cite{all_freq} related to the spherical signed distance (SSD) $\theta$ and the sharpness $\lambda$ of an SG. For the denominator, it can be quickly evaluated in a manner similar to~\cref{SG_render_eq}.
%we perform SG inner product similar to~\cref{SG_render_eq}.

%multiply the SG amplitude of the incident light by the ratios $\gamma$ and render the virtual object using the updated incident light. We define $\gamma$ as the ratio of the SG integral of the unobstructed region to the integral of the full hemispherical region, which is given by:

For adding the self-shadows, we first attenuate each SG for the incident lighting with a factor $\gamma_k$, and then render the virtual object. $\gamma_k$ is the ratio of the integral of the unobstructed region to that of the complete upper hemisphere:
\begin{equation}
\gamma_k(\mathbf{x})=\frac{\boldsymbol{\mu}_k f_h(S(\mathbf{p}_k; \mathbf{x}), \lambda)}{\boldsymbol{\mu}_k f_h(\frac{\pi}{2}, \lambda)}=\frac{f_h(S(\mathbf{p}_k; \mathbf{x}), \lambda)}{f_h(\frac{\pi}{2}, \lambda)}.
\label{self shadow ratio}
\end{equation}
Note that here we do not attenuate with $\kappa$ to avoid computing the shadows twice.

\begin{figure*}[!ht]
\centering
\includegraphics[width=1.0\textwidth]{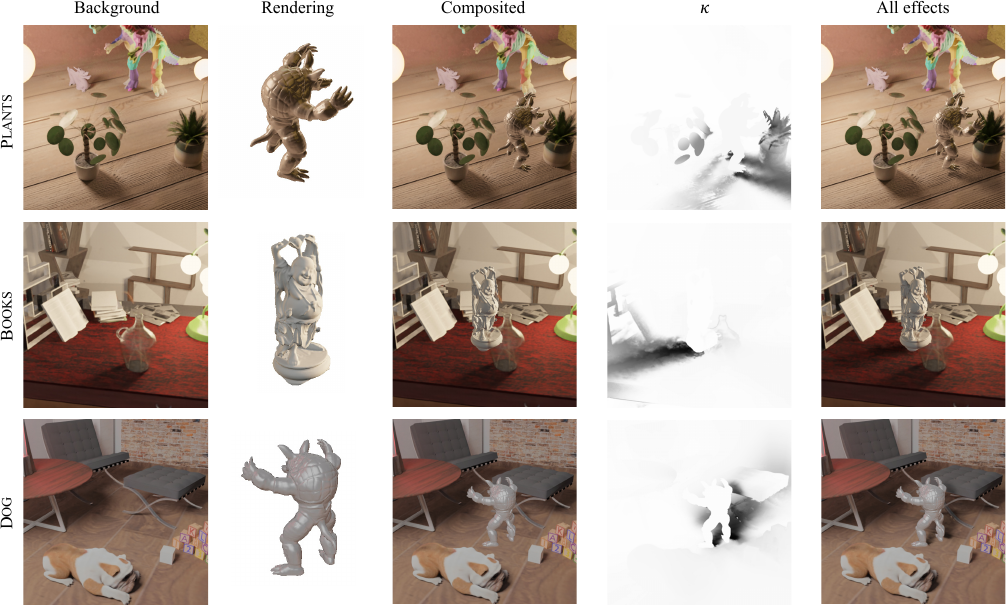}
\caption{Visualization of intermediate and final object insertion results of the \textsc{Plants}, \textsc{Books} and \textsc{Dog} synthetic scenes. From left column to right, background images, virtual objects rendering results, compositing results with occlusion or translucency effects, $\kappa$ (\cref{shadow ratio}), object insertion results with all effects.}
\label{intermediate}
\end{figure*}

\begin{figure*}[!ht]
\centering
\includegraphics[width=1.0\textwidth]{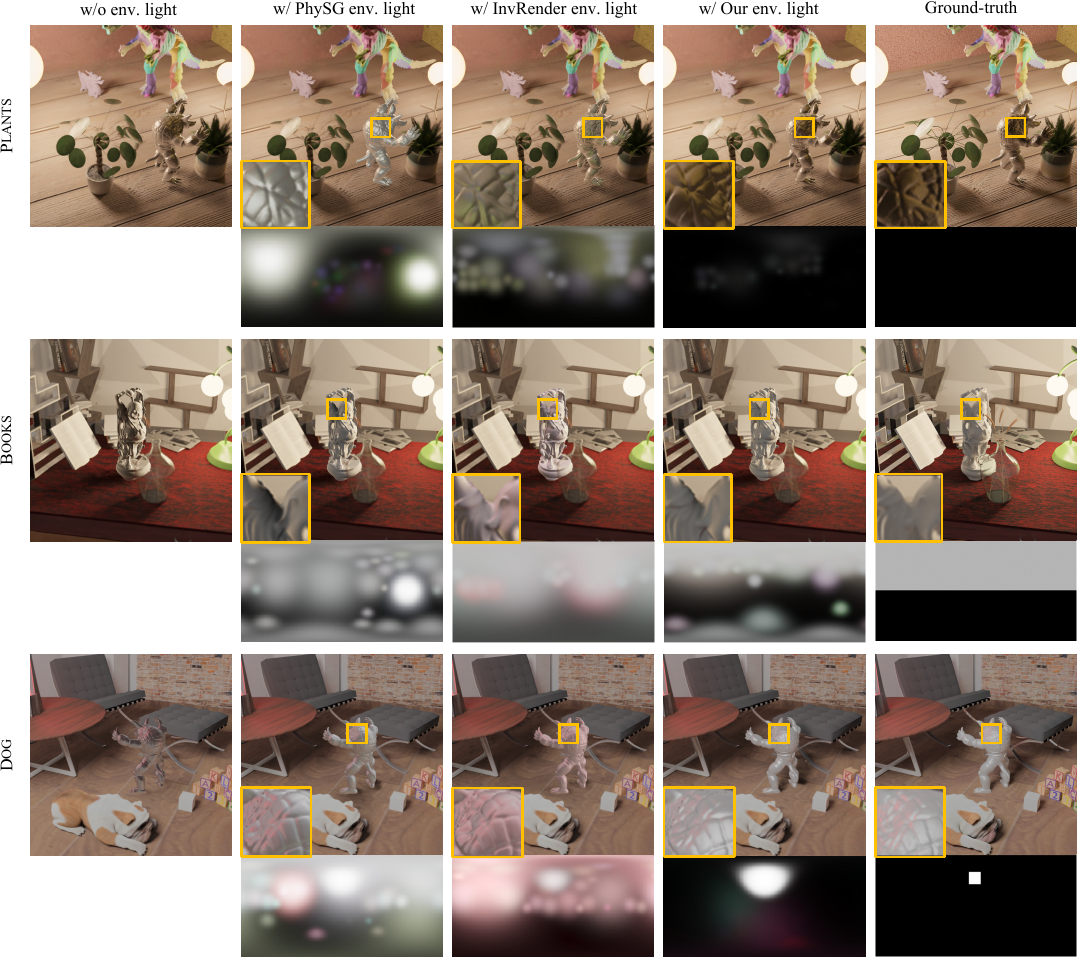}
\caption{A qualitative comparison of using different environment lighting in \textsc{Plants}, \textsc{Books} and \textsc{Dog} synthetic scenes. In the first column, virtual objects are rendered by only using the NeRF near-field lighting. From the second left column to the fourth column, we compensate different estimated environment lighting estimated by the method of PhySG~\cite{physg}, InvRender~\cite{zhang2022modeling} and our proposed method. The last column is the ground-truth rendered by Blender Cycles. Some details are magnified for better comparison. The environment lighting is shown by equirectangular projection under the rendered images.}
\label{woenv}
\end{figure*}

\begin{figure*}
\centering
\includegraphics[width=1.0\textwidth]{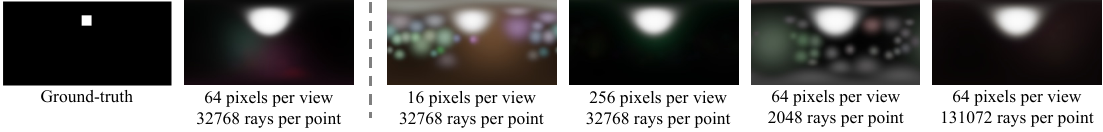}
\caption{Estimating environment lighting by different sampling strategies. All the results are estimated in \textsc{Dog}. The result of our proposed method is shown on the second from left, which samples 64 pixels per view and traces 32768 rays at each surface point corresponding to the pixel. Results estimated by other sampling strategies are shown on the right side of the dashed line.}
\label{rays_pixels}
\end{figure*}

\begin{figure*}[ht]
\centering
\includegraphics[width=1.0\textwidth]{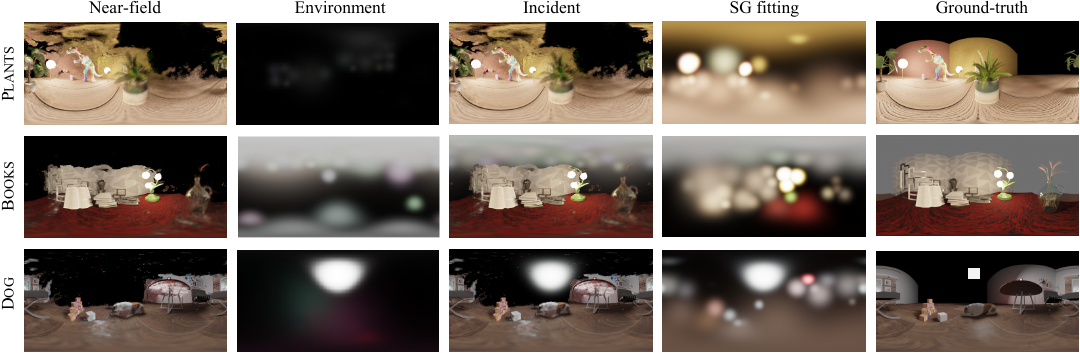}
\caption{Visualization of the lighting at the virtual object center. From left column to right, near-field lighting, environment lighting, incident lighting, SG fitted incident lighting and the ground-truth lighting rendered by Blender Cycles. The incident lighting is the blend of the near-field and environment lighting according to the NeRF opacity.}
\label{incident}
\end{figure*}

\section{Results}
\label{experiment sec}
All experiments are conducted on a workstation with an Xeon(R) Platinum 8352V CPU, 90GB memory and an NVIDIA RTX 4090 GPU. Our pipeline achieves a performance of 18 frames per second on average at a resolution of $1280\times720$. \warning{Specifically, it takes about 40ms for instant-ngp to render the NeRF. And it takes 15ms for virtual object insertion, including 4ms for incident lighting computation, 7ms for SG updating (except that 84ms is needed for the first frame), and 4ms for the rendering, compositing and shadow casting of the virtual object. The average GPU usage is 55\%, which shows room for future improvement over memory access.}
%which indicates that our method still has the potential for further optimization in implementation to break memory bottlenecks.

%We implement an editor with interactive GUI that supports to place a virtual object, edit its material and observe the scene from free views. 
%The editor will pass required parameters to the rendering pipeline and present the rendering result in real time. We implement our rendering pipeline based on ngp-pl
%\footnote{ngp-pl. \href{https://github.com/kwea123/ngp_pl}{https://github.com/kwea123/ngp\_pl}}
%
%, which is an unofficial Pytorch implementation of instant-ngp. Compared to the original implementation, it runs relatively slow but is easy-developing. 
%Running on a NVIDIA RTX 3070 GPU, our system can render images of $800\times800$ at an average of 14 fps. 
For precomputation, it takes about 25 minutes to estimate the environment lighting from an input NeRF. For a virtual object, we spend 8 minutes to precompute the SSDF field, which takes up 9MB. Note that the SSDF field can be used with any NeRF.

%including 15 minutes for high-quality NeRF training and 30 min for environment lighting estimation. For each virtual object, it takes about with 9 MB storage consumption. Once the precomputation is completed, the virtual object can be used in any scene. 

In the following, we describe the data related to our experiments, including synthetic and captured HDR NeRFs, in~\cref{datasets}. Ablation studies are demonstrated in~\cref{ablation}. Finally, we compare against state-of-the-art methods  in~\cref{comparisons}.

%multi-view datasets to test the proposed method  Then, we perform ablation studies on synthetic datasets to discuss our key components in~\cref{ablation}. Finally, we make qualitatively comparisons with state-of-the-art methods to show our realistic object insertion results in~\cref{comparisons}. For all HDR image results, we perform tonemapping using the method of Reinhard et al.~\cite{reinhard_tonemapping} with the same parameters to display LDR image in the following experiments.

\subsection{Data}
\label{datasets}

We build four synthetic HDR NeRFs by training using images rendered from multiple views with Blender Cycles~\cite{BlenderCycles}.
%\footnote{Blender. Version 3.4}%
%We use the most prominent object in each scene to name it. 
For \textsc{Plants}, all light sources can be observed in the rendered images, and therefore the environment lighting should be totally black. For the other 3 NeRFs (\textsc{Books}, \textsc{Dog} and \textsc{Cups}), there are some light sources that do not appear in any of the rendered images, which are considered as the environment lighting.
%We use these scenes to verify the necessity of the global light estimation in the case of missing complete lighting information. 
%The reason we use synthetic scenes is to conveniently obtain the ground-truth images by placing the virtual object in the scene with the same parameters from the editor and rendering it with Blender. 

In addition to synthetic NeRFs, we also construct two NeRFs by training with captured HDR photographs from two real scenes (\textsc{Horse} and \textsc{Sheep}). We set the camera to the bracketing mode with the step of 1.5 EV and shoot 9 images with different exposures for each camera pose. Each HDR image is recovered from 9 LDR images using~\cite{debevec_hdr_recover}. 

To provide the ground-truth for object insertion experiments, we 3D-print several models and uniform sprayed them with paints to simulate different materials. We apply Ma et al.~\cite{ma2021free} to measure the material parameters, including roughness and metallic, and use the same parameters for virtual objects rendering. The 3D printed models are placed in the real scene, and images are taken around them as the ground-truth. We use the images without the presence of the model to train NeRFs, and compare object insertion results with the images with the physical model in place. The sizes of virtual objects are adjusted manually to ensure that it is closed to the ground-truth. 
%All related images are cropped to a resolution of $800\times800$.

%ground-truth images from bothsynthetic and real scenes are used for reference only. Our methods always provides biased results for some reason. For example, Blender render objects using Monte Carlo path tracing to precisely solve the render equation offline while our SG-based renderer simplifies the render equation for real-time rendering. In addition, the material of real objects also differs from the standard Cook-Torrance model. Effects like subsurface or anisotropic reflection are not considered in our rendering method. Therefore, the following experiments mainly perform qualitative comparison and analysis. All of the outputs in experiments are HDR images, so we perform tonemapping using the method of Reinhard et al.~\cite{reinhard_tonemapping} with the same parameters for each output to present results.

%due to the use of SG rendering method,

\begin{figure*}[h]
\centering
\includegraphics[width=1.0\textwidth]{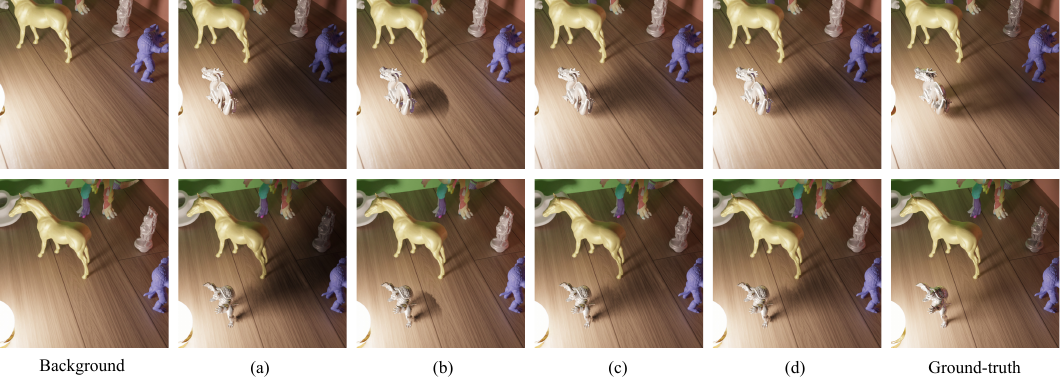}
\caption{Comparison of shadows generated by using the $16\times16\times16$ SSDF field that obtains SSD values at far positions by clamping to the boundary values (a), truncating to positive infinity (b) and approximating from the boundary values (c), and shadows generated by using the $64\times64\times64$ SSDF field that interpolates SSDF at any position (d). The ground-truth is rendered by Blender Cycles.}
\label{ssdf cmp}
\end{figure*}

\subsection{Ablation studies}
\label{ablation}
We first perform ablation experiments to show the intermediate results of each step described in~\cref{object insertion sec}. In~\cref{intermediate}, the virtual object shows all-frequency effects with realistic reflections.
We can see the virtual object through the glass bottle in the \textsc{Books} scene, which is a challenging task for traditional AR methods~\cite{macedo2021occlusion}. The shadows computed by our method also considerably enhanced the visual realism.
%Shadow effects convince the presence of the virtual object, which is physics-based without the limit on the number of light sources. All the above effects make it difficult to distinguish the inserted object from the surrounding environments.
%We create occlusion or translucency effects by changing the sampling range of NeRF and performing blending between the rendered object and the scene, which makes 
%We also prove the necessity of the compensated environment lighting and the impact of modeling local lighting for environment lighting estimation in~\cref{woenv}.
%The main difference between the three methods is how to model the lighting.
%supposes that the environment light is the full incident light, so it does not take the NeRF near-field lighting and visibility into consideration.

Next, we show the impact of our estimated environment lighting as well as the near-field lighting represented in the NeRF in~\cref{woenv}. Without the environment lighting which estimates the sources not covered in the NeRF, the virtual object appears darker in the \textsc{Books} and \textsc{Dog} scene. Moreover, we compare our method with PhySG~\cite{physg} and InvRender~\cite{zhang2022modeling} on environment lighting estimation. These three methods have a similar estimation framework, and all represent the environment lighting as 32 SGs in our experiments.  PhySG does not take into account the near-field lighting and visibility.  InvRender roughly models the near-field lighting by training a MLP to cache ray-tracing results. Our method more accurately model the near-field lighting (especially the visibility, cached to texture maps), and sparsely sample the surface points to reduce memory consumption. In the \textsc{Plants} scene, all light sources are included in the NeRF, and therefore the ground-truth environment lighting is totally black. PhySG treats the light sources in NeRF as the environment lighting. The virtual object is brighter than the ground-truth in the \textsc{Plants} scene, while the self-shadow is too dark in the \textsc{Books} scene. For InvRender, caching near-filed lighting with MLP by tracing 16 rays at each point is too rough for complex scene-level NeRFs, which leaves a lot of ambiguity during optimization. Our method provides the best results, and the estimation is not affected by the light sources incorporated in the NeRF.

%In addition to validating the necessity of the local lighting, 
In addition, we evaluate the impact of the number of pixels and rays over the estimated environment lighting in~\cref{rays_pixels}. If we sample less pixels, the estimation will suffer from the ambiguity of albedo and shading. If we trace less rays at each point, the near-field lighting is less accurate, resulting in a highly noisy estimate. In comparison, sampling fewer pixels has a greater impact than sampling fewer rays. Sampling more pixels and rays usually leads to slightly better results, but bears the increasing costs of time and memory.

To validate the our lighting model in~\cref{incident_light}, we show the near-field lighting (from NeRF), estimated environment lighting, full incident lighting (before and after SG fitting), and the ground-truth in~\cref{incident}. Our method generates high-quality incident lighting close to the ground-truth. After fitting with SGs, the main light sources are preserved. We notice that there are some extra floaters (\textsc{Plants} scene) and missing parts (\textsc{Books} scene) in the near-field lighting. Nevertheless, they do not notably influence the rendering as the brightness is relatively low. Also, the estimated environment lighting well compensates for the lighting not covered by NeRF.
%After adding the global light, the virtual object appears more reasonable and casts shadows to the same direction as other objects in the scene. Although our global light estimation is low-frequency, with the supplementary details of the indirect light, the insertion results are still convincing.
% replace the first row of our results, it is too bright

% In Section~\ref{NeRF train}, we propose using the depth loss to reduce floaters near cameras. As shown in Figure~\ref{fig_depthloss_cmp}, scenes reconstructed by NeRF without the depth loss are over-fitted. Floaters or tiny image planes fill in the details of scenes to better fit the target images when training, which not only leads to messy geometry, but also reduces the quality of novel view synthesis. The reasonable results are generated after adding the depth loss.

% \begin{figure}[h]
% \centering
% \includegraphics[width=0.4\textwidth]{imgs/第2页.pdf}
% \caption{Comparison of scenes reconstructed with or without the depth loss.}
% \label{fig_depthloss_cmp}
% \end{figure}

To validate our method for extrapolating SSDF (\cref{SSDF fields sec}), we compare in~\cref{ssdf cmp} the shadows generated by our parameters and using a $64\times64\times64$ SSDF field with an extended range of $-6r_\mathrm{obj}$ to $6r_\mathrm{obj}$, where $r_\mathrm{obj}$ is the radius of the virtual object bounding sphere, so that the SSDF is interpolated at any pixel in the rendered image. There are no significant differences between the results computed with two sets of parameters, while the simple clamping method generates unsatisfactory results in regions far away from the virtual object. Compared to the ground-truth rendering results, our shadows appear slightly darker, due to the approximation of the inner product of the environment lighting and the visibility term.

%In~\cref{SSDF fields sec}, we approximate the spherical signed distance (SSD) value at the position far away from the virtual object by the SSD value at the position closed to the object to reduce precomputation. We compare the shadows generated by using the $16\times16\times16$ SSDF field with the range of $-1.5r_\mathrm{obj}$ to $1.5r_\mathrm{obj}$ that obtaining SSD values at far positions by approximating from the boundary value, clamping to the boundary value and truncating to infinity, and shadows generated by using the  $64\times64\times64$ SSDF field with the range of $-6r_\mathrm{obj}$ to $6r_\mathrm{obj}$ that using the interpolated SSDF at any position. The comparison results are shown in~\cref{ssdf cmp}. 

\begin{figure*}[!ht]
\centering
\includegraphics[width=1.0\textwidth]{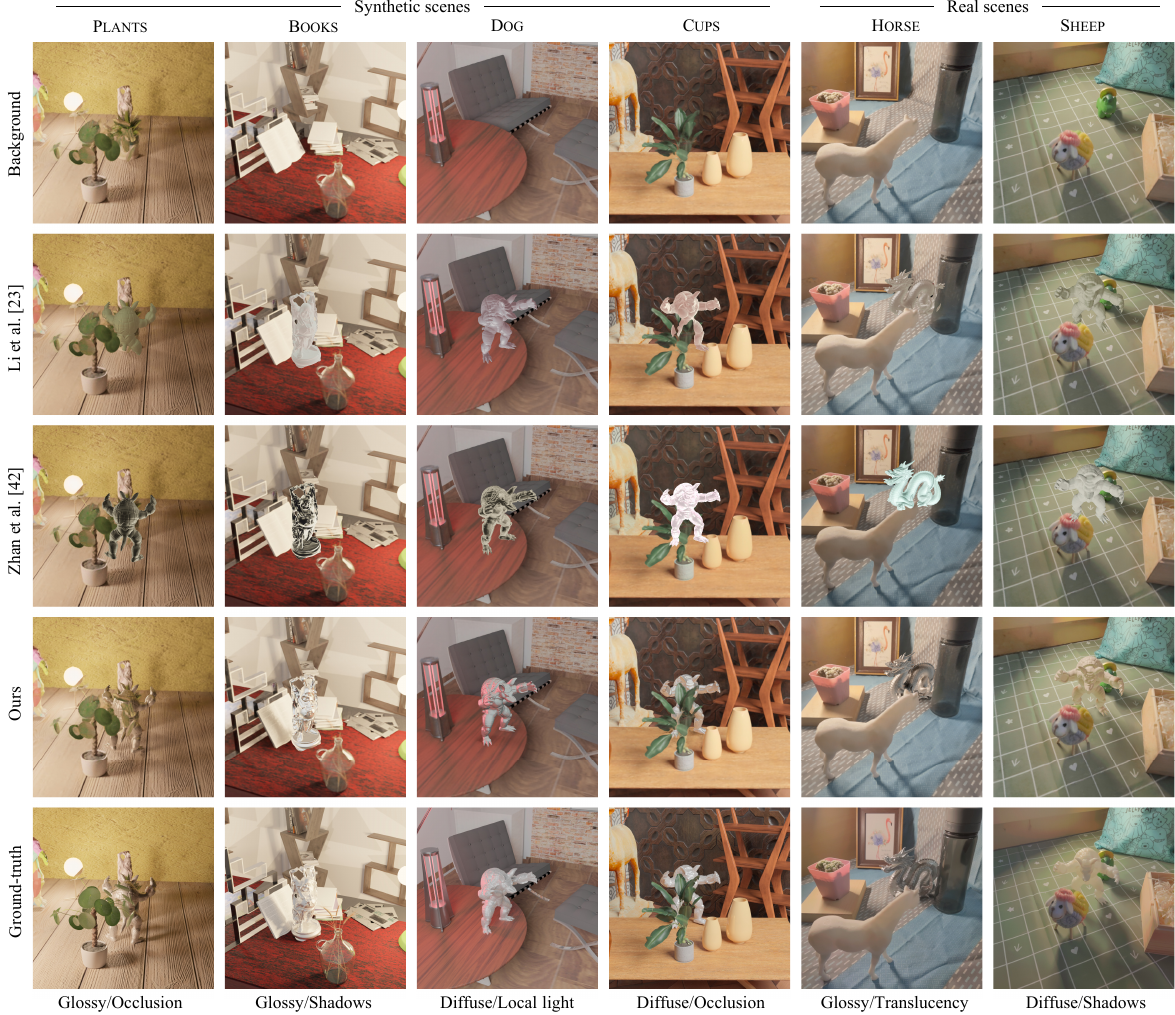}
\caption{Comparison with the work of Li et el.~\cite{IRAdobe} and Zhan et al.~\cite{emlight} in both synthetic scenes and real scenes. The materials of virtual objects and the noteworthy effects are indicated under each column.}
\label{fig_cmp_res}
\end{figure*}

\subsection{Comparisons}
\label{comparisons}
We first compare our approach with two state-of-the-art methods, Li et al.~\cite{IRAdobe} and Zhan et al.~\cite{emlight}. Both take RGB images as input and employ neural networks to estimate the lighting. The former outputs 2D spatially-varying incident lighting, from a joint optimization of material parameters and depths. And the latter only outputs one incident lighting. For comparison, we first render (for synthetic scenes) or capture (for real scenes) images of a scene without the virtual object, and use them as input for both methods. The estimated lighting are represented as SGs, so that we can render the virtual object as described in~\cref{object insertion sec} to compare across different methods. Depth test is performed in~\cite{IRAdobe} to account for occlusions. Since the method of~\cite{emlight} does not provide any geometric information, we directly place the virtual object on top of the background. We do not compare our shadows with the baseline methods, as both of them generate shadows on manually specified planes. Instead, we compare our shadows with the ground-truth. 

%For shadows, both methods use off-the-shelf renderers to generate shadows on manually specified planes, which is not suitable for comparison with our method. Therefore, there are no shadows in the results of the two methods. We mainly compare shadow effects with the ground-truth. 
%The comparison results of synthetic and real scenes are shown in~\cref{fig_cmp_res}. 
%The comparison results of synthetic scenes are shown in Figure~\ref{fig5} and real scenes are shown in Figure~\ref{fig_real}. 
%To compare noteworthy details like shading, shadow and occlusion effects, we use boxes of different colors to frame them and present the corresponding enlarged image. %due to their weak generalization ability
\cref{fig_cmp_res} shows the comparisons on synthetic and real scenes. Virtual objects rendered with~\cite{IRAdobe} or~\cite{emlight} can be easily distinguished from the background. These two methods are trained with indoor image datasets and are likely to estimate results with considerable bias. In comparison, our method is able to robustly estimate the incident lighting. Our rendering results exhibit realistic reflections and shadows that closely resemble the ground-truth, regardless of the material, e.g. diffuse in \textsc{Dog} or glossy in \textsc{Books}. Even for scenes with complex lighting, such as multiple light sources in \textsc{Plants} and the environment lighting with local lights in \textsc{Dog}, our method is able to produce reliable results. The depth map computed by Li et al. is rough, leading to noticeable cracks or overlaps, e.g., in \textsc{Plants} and \textsc{Cups}. Note that our method supports the transparency and translucency represented in the input NeRF (see the plastic cup in \textsc{Horse} in~\cref{fig_cmp_res} and the glass vase in \textsc{Books} in~\cref{intermediate}).

\begin{figure}
\centering
\includegraphics[width=0.47\textwidth]{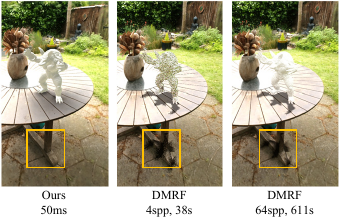}
\caption{Comparison with DMRF~\cite{qiao2023dynamic}. It takes 50ms to render the frame by our methods while DMRF takes 38s to render a 4spp frame and 611s to render a 64spp frame. Leaked shadows are observed with their approach.} 
\label{DMRF}
\end{figure}

\warning{In~\cref{DMRF}, we further compare with the concurrent work of DMRF~\cite{qiao2023dynamic}. There are 3 major differences. First, based on a path-tracing framework, their performance cannot reach real time. Second, their approach does not estimate light sources not covered the input NeRF. Finally, the visibility in the NeRF is ignored, leading to artifacts like shadow leakage, as shown in~\cref{DMRF}.}

\begin{figure}
\centering
\includegraphics[width=0.47\textwidth]{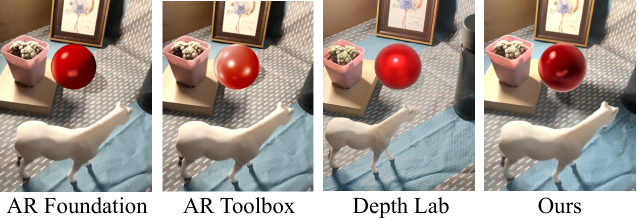}
\caption{Comparison with popular AR frameworks in industry, including Unity AR Foundation~\cite{AR_Foundation}, AR Toolbox~\cite{AR_Toolbox} and Google Depth Lab~\cite{depthlab}.} 
\label{fig6}
\end{figure}

Moreover, we qualitatively compare against popular AR frameworks from the industry, including Google Depth Lab~\cite{depthlab}, AR Toolbox~\cite{AR_Toolbox} and Unity AR Foundation~\cite{AR_Foundation}, due to the difficulty for precisely controlled experiments. As the common types of virtual object supported by all frameworks are limited, 
we only test a red ball in the \textsc{Horse} scene. As shown in~\cref{fig6}, our method produces more accurate and detailed rendering with soft shadows, while the counterparts only implement hard shadows on estimated planes.

\subsection{Limitations} 
\begin{figure}[h]
\centering
\includegraphics[width=0.47\textwidth]{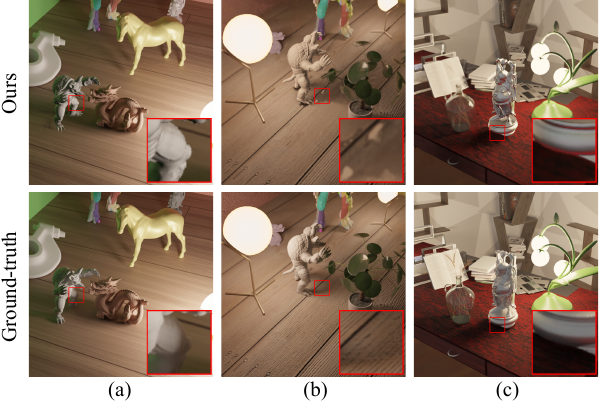}
\caption{Failure cases: (a) shadow missing due to the assumption that all surface points of the virtual object shares the same incident lighting; (b) noisy shadow caused by inaccurate depth estimation; (c) incorrect shading due to only considering the direct illumination.}
\label{fig_fail}
\end{figure}

First, as shown in~\cref{fig_fail}(a), certain local lighting effects may not be modeled, due to our assumption that all surface points of the virtual object shares the same incident lighting. Second, due to the inaccurate depth estimation in NeRF, there is a small amount of noise in our shadows in~\cref{fig_fail}(b). Finally, we consider direct illumination only, which may lead to artifacts. For example, in~\cref{fig_fail}(c), the virtual object rendered by our method shows red reflections from the wood table. However, this is incorrect, as this part of the wood table is actually occluded by shadows of the object.
%our method only use the incident light at the object center to render the object. When the incident light significantly changes on the virtual object surface, some local effects can not be modeled. \cref{fig_fail}(b) presents noisy shadows caused by inaccurate surface points estimated by NeRF. In~\cref{fig_fail}(c), the virtual object rendered by our method reflects red color from the wood table, but this part of the wood table is actually covered by shadows, which should be black as the ground-truth. Besides, our method only considers the direct illumination and shadows, but in fact, there are complex multi-bounce light effects, e.g. the reflection of the virtual object on the smooth surface in the NeRF. These effects are subtle but require a lot of computation. Previous work like~\cite{pan2007precomputed} attempts to achieve interreflections by huge precomputations, which is impractical for application. We look forward to better solutions in future work.  

%For specular or textureless surfaces, NeRF may cause reflections below the surface or messy floaters in the space, which affects our shadows that are based on the surface estimation.
%Our shadow method depends on robust and reliable geometry information estimated from NeRF, which needs to be improved in future work.

\section{Conclusion}
\label{conclusion sec}
We propose a real-time method for inserting virtual objects into a NeRF. Our key contribution is the first framework that fully exploits the lighting and geometry information in NeRFs to enable realistic rendering with occlusion and shadowing effects. We outperform state-of-the-art techniques in terms of quality. Compared with existing work, combining NeRF with AR has considerable benefits and may be widely deployed with the popularity of NeRF. We hope that this paper could inspire further research along this direction.

{\small
\bibliographystyle{cvm}
\bibliography{myBib}
}

\end{document}